# A Factorized Recurrent Neural Network based architecture for medium to large vocabulary Language Modelling

Anantharaman Palacode Narayana Iyer
JNResearch, Bangalore, India
ananth@jnresearch.com

*Abstract*— **Statistical language models are central to many applications that use semantics. Recurrent Neural Networks (RNN) are known to produce state of the art results for language modelling, outperforming their traditional n-gram counterparts in many cases. To generate a probability distribution across a vocabulary, these models require a softmax output layer that linearly increases in size with the size of the vocabulary. Large vocabularies need a commensurately large softmax layer and training them on typical laptops/PCs requires significant time and machine resources. In this paper we present a new technique for implementing RNN based large vocabulary language models that substantially speeds up computation while optimally using the limited memory resources. Our technique, while building on the notion of factorizing the output layer by having multiple output layers, improves on the earlier work by substantially optimizing on the individual output layer size and also eliminating the need for a multistep prediction process.**

*Keywords-Recurrent Neural Networks; Language Models; hierarchical softmax; class based prediction*

## I. Introduction

Language Models are widely used in several natural language processing tasks such as Information Retrieval, Machine Translation, Speech Recognition etc. Language models assign a probability to an input text and this problem can also be formulated as a sequential data prediction where the model predicts the next word given the previous words of a text. While the traditional n-gram language models can only handle limited contexts without blowing up the size of model parameters, connectionist approaches, such as convolutional neural networks and recurrent neural networks have shown a lot of promise towards handling larger contexts. Recurrent Neural Network (RNN) based architectures, with their ability to support an arbitrarily sized context have been reported to outperform most of the state of the art n-gram systems [1]. Mikolov et al. report a 50% reduction in perplexity using a mixture of RNNs compared to the state of the art traditional models [2]. RNNs, like other neural network based language models, predict the next word of a sequence by computing the probability distribution over the entire vocabulary using a softmax layer. Each unit in the softmax output layer represents a unique word in the vocabulary. An increase in the size of the vocabulary results in a corresponding increase in the size of the output layer. Thus, computing softmax distributions for a large vocabulary becomes computationally expensive as the model parameters, specifically the weight matrix between the output and hidden layers becomes larger.

The motivation for our work emanated from the need to train and deploy language models involving large vocabularies on typical off the shelf computing devices using the popular libraries supported on Python. Our goals are to develop an architecture that enables: (a) Reduced training and prediction time for vocabularies with size > 10000 words (b) Meeting or exceeding the model performance compared to other RNN based architectures (c) A flexible architecture that scales well with the data size.

The language model reported in our paper is targeted at supporting medium to large size vocabularies with vocabulary sizes in the range 10000 to 50000 words. Several approaches have been proposed and evaluated in the past that aim at reducing the computational complexity by reducing the number of model parameters. Some of the approaches are aimed at modelling very large corpora with hundreds of millions of word tokens [5]. The recent advances such as the class based factorization approaches [3][14] demonstrate the feasibility of improving the training speed without significant degradation in performance compared to full softmax. Morin and Bengio [7] first introduced the hierarchical softmax technique, that was improved upon subsequently in the work reported in [4][6]. These techniques where the output words become the leaf nodes and the internal nodes defining the relative probabilities of their child nodes, reduce the computation to log N complexity instead of evaluating N output nodes. Another recent approach by Huang et al. [8] that is also based on an RNN architecture for language modelling, uses a two-level hierarchy of classes where words are binned in to classes by word frequency and classes are categorized in to super classes according to the class frequency. Our architecture, while leveraging the notion of output layer factorization, takes a different approach that obviates the need for a multi-step hierarchical prediction presented in the recent literature while achieving comparable computation efficiency. We partition the input space in to equivalence classes that share an output layer in an optimal manner. In this paper we describe our model and show that the results are better than or comparable to the other RNN based systems. Our technique, that factorizes the output layer in a novel way, achieves a speed up by a factor of around 32 on Brown corpus as compared to the RNN with full softmax layer approach.

## II. RELATED WORK

Bengio proposed a feed forward artificial neural network model that takes as input a fixed length context and generates a softmax probability distribution over the vocabulary. The network also learns the input word representation simultaneously with the language model [9]. One advantage of this model is that as the word representations are learnt by the network, the n-grams not observed in the training corpus can be handled effectively. However the fixed size context, a large softmax size and the simultaneous learning of an accurate word representation are major deficiencies [2]. Collobert and Weston use convolution based networks for multitask learning and predict the outputs for six core NLP tasks: Part-of-speech (POS) tagging, Chunking, Named Entity Recognition (NER), Semantic Role Labeling (SRL), Language Models and Semantically Related Words [10][11]. Convolution based neural network approaches also are required to work with a fixed size context window. As the RNNs can handle variable length contexts, intuitively, they are a natural fit for language model predictions where there could be long range dependencies between the words. In their paper [2] Mikolov et al. reported a language model based on a simple recurrent neural network or Elman network architecture [13], which is easy to implement and train. In order to improve the efficiencies, the model described in [2] was extended in to a RNN with the single large output layer factorized in to multiple smaller layers with the addition of a class layer. This model attempts to predict the probability of the next word given the context by first predicting the class to which the next word belongs and then predicting the probability distribution of the words within the predicted class (Fig 1). This model has the advantage of reducing the computations substantially as only a smaller subset of the output vocabulary are considered for generating the probability distribution as compared to the full softmax. Concretely, the time complexity of a training step for the full (without factorization) softmax layer is proportional to:

$$O = (1 + H) \times H \times \tau + H \times V \quad (1)$$

where: H is the number of hidden units of the RNN, τ denotes the time steps through which we backpropagate, as per the back propagation through time (BPTT) algorithm [12] and V is the size of the output layer that is also the size of the vocabulary for the full softmax prediction. When V >> H, which is usually the case, from the equation (1) we observe that the size of the output layer causes the computational bottleneck due to $H \times V$ term. With the factorization approach the equation (1) reduces to:

$$O = (1 + H) \times H \times \tau + H \times C \quad (2)$$

where C is the number of classes. By setting a value of C to be a small fraction of V, it is possible to substantially speed up the training time. As we observe, this model requires a separate class layer and the network needs to predict the class and the distribution within the class. The assignment of words to classes is done in accordance with their unigram distributions. While this approach provides the probability distribution of words within a class c, many words in c may not be the possible words that can follow the input context regardless of their unigram probability.

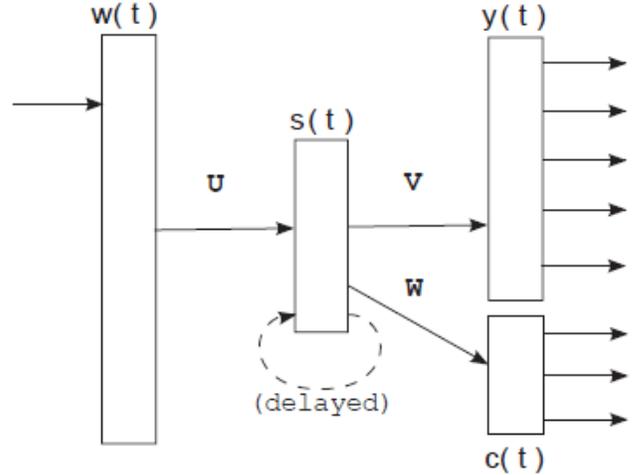

**Fig 1**: RNN with output layer factorized by the class layer (ref Mikolov et al.)

As an example consider the sentence: "The Microsoft Lumia 550 runs Windows 10 Mobile and is powered by a 2100mAh removable battery". Suppose in the training corpus Windows 8 occurs a large number of times compared to Windows 10, as Windows 10 is a more recent version. This scenario might result in larger unigram counts for 8 and much less for 10, causing these versions to be categorized in to different classes. If this happens, the language model will not consider Windows 10 as a probable sequence, causing inaccuracies when applied to real world problems, as the desirable probability distribution would have been across version numbers like 7, 8, 10 etc. Furthermore, the accuracy of the predicted distribution critically depends on the classification accuracy for the class c in C where C is the set of classes.

We present an RNN based architecture that builds on the factorization approach but alleviates the aforementioned drawbacks of earlier approaches, particularly, the multistep prediction process and the unigram counts based class partitioning.

## III. MODEL DESCRIPTION

### A. Rationale

A full softmax classifier for language modelling generates a probability distribution across all words of the output vocabulary, given the context words. However, for most real world applications, this might be an overkill as not every word in the vocabulary may have a reasonable probability of following every other word. We propose an architecture that leverages this observation, that is, any word w in the training corpus may be followed by only a limited set of words in the vocabulary V that is often a much smaller subset of V. This list that we may also denote as the "follow" list, is given by the bigram distribution of the word. Suppose B(w) is the list of bigram keys of the word w, we observe: |B(w)| << |V|. This is shown in Fig (2) for the Brown corpus and Fig (3) for our custom corpus that has text on product reviews for mobile devices, where the labels Single, Tiny, Small, Medium, Large, XLarge, XXLarge, Ultra denote the non-overlapping frequency intervals (bins). In our experiments, these data labels correspond to the size thresholds: 1, 32, 64, 128, 256, 512, 1024 and above 1024 respectively.

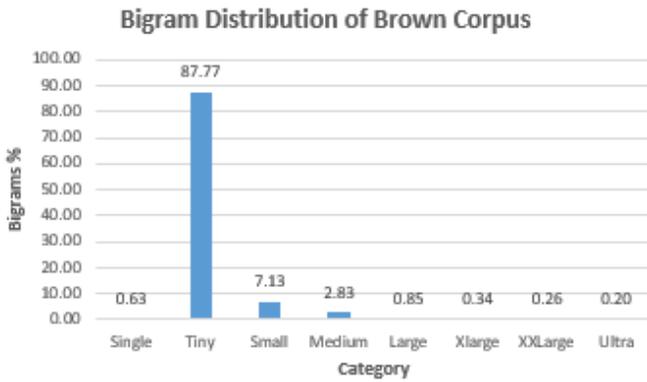

**Fig** 2: Bigram distribution of Brown corpus (NLTK 3.0 distribution)

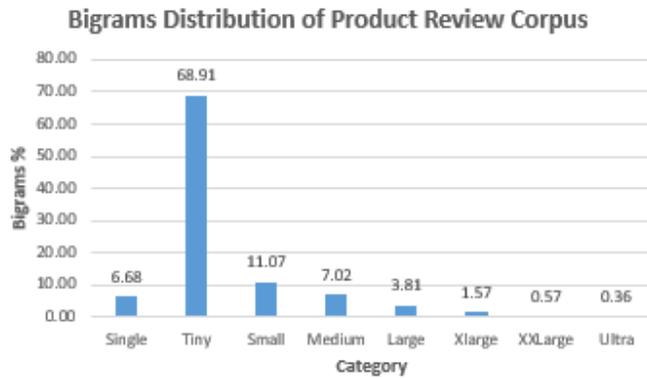

**Fig** 3: Bigram distribution of Product Review custom corpus

As we observe from the distributions shown in the figures, the bin that represents Tiny size category has the largest number of bigrams. This suggests that most words in the vocabulary have bigrams whose lengths are upper bounded by 32. Such a pattern could be typical for many corpora that follow an 80/20 rule of distribution and hence is a fairly general characteristic.

From this we may conclude that an architecture with output layers having a variable number of output units would optimize the number of hidden to output layer computations.

*B. Architecture*

Our system architecture consists of an RNN with a single hidden layer and a number of output layers, each output layer constituted by a subset of the vocabulary. This is a factorized output layer model without the additional class layer (Fig 4). A sentence of n words is represented as an n element sequence, where each element at t is a word vector $x_t$ corresponding to the word $w_t$. This constitutes the input layer of the RNN at that time instant. The hidden layer at a time t receives its inputs from the current input layer $x_t$ and also from the hidden activations of the previous time step $h_{t-1}$. In our model, each word in the input vocabulary is assigned its own dedicated output layer and multiple words may share the same output layer.

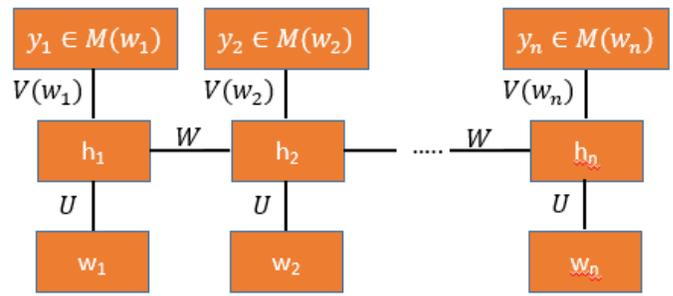

**Fig 4**: RNN model where the output $y(t) \in M(w_t)$ and M is a function that maps an input word to its corresponding output layer. V(w) maps the word w to its corresponding weights for the hidden to the output layer. U and W are the respective input to hidden and hidden to hidden weight matrices.

The activations are computed as follows:
$$h_t = f(Ux_t + Wh_{t-1} + b_h) \quad (3)$$
$$y_t = softmax(V_t h_t + b_{yt}) \quad (4)$$
$$P(w_t|w_1, w_2 \dots w_{t-1}) = g(y_t, w_{t-1}) \quad (5)$$
where: g is a function that maps the current output from the softmax layer to the probability of word w in the output vocabulary. It may be noted that the output vocabulary could be different for each word in the input vocabulary and in almost all cases, it is a small subset of the vocabulary generated from the corpus. The output layer for a word w is determined by the function V(w). Each step t in the RNN takes an input word $w_t$ in its vectorized representation. We denote its corresponding output to hidden layer weight matrix as $V_t$. Thus, $V_t = V(w_t)$. Similarly, the bias terms for the output layer also is a function of the word $w_t$ at step t.

As our model factorizes the full softmax in to a set of output layers, the computation of the output activations proceeds by:
a) Determining the output layer that corresponds to the given input
b) Determining the weight matrix $V_t$ and the bias vector $b_{yt}$ to be used for the given input.
c) Computing the output activations using the selected output layer parameters $V_t$ and $b_{yt}$
d) Assigning the probability distribution computed as above to the words that constitute the output layer.

We make the following observations on our model:
- An input word has a unique 1 to 1 mapping with an output layer. That is, each input word is assigned to exactly one output layer.
- The words in one output layer may occur in any other. That is, the output vocabularies of different output layers may have overlaps.
- Many input words may be mapped to a single output layer. More the number of words that get mapped to a single output layer, less will be the number of such layers to be generated.

The critical step in achieving the model efficiency and performance is centered around finding the optimal output layer assignment for each word in the input. This is a preprocessing step before the RNN is trained. The output layer assignment specifies the number of softmax units and a mapping of each unit to its corresponding word in the vocabulary.

*C. Output Layer Assignment*

We need to determine the output layers that require minimal computation for a given input. It is wasteful to evaluate a large number of softmax units for an input that has only a very small number of words that can follow it. This suggests that assigning a dedicated output layer with the number of output units same as the number of words in the bigrams of the input is the optimal fit. However there are a couple of issues with this approach. Firstly, this requires us to have as many output layers as the size of the vocabulary. Secondly, by mapping an output layer for the word w exactly to be its bigram words B(w), the ability of the model to generalize for words that are not in B(w) but are part of the vocabulary is hampered. Our procedure to determine the output layers is as below.

We start by using the "follow" list of each word, given by its bigram, to determine the vocabulary of the output layer ($V_o$) as seen in the training corpus. The intuition behind this is that the bulk of the probability mass for the distribution we are predicting should be concentrated around the bigram of the current word, regardless of the context. Thus, bigrams provide the starting point for assigning the words to an output layer. But as each word in the input has its own follow list, mapping each follow list to a dedicated output layer is not efficient as we need to train and manage as many output layers as the size of the vocabulary. However, this scenario is the worst case where we do not take in to consideration the intersections between the bigrams of input words. For instance, the words "the" and "a" may have bigrams B("the"), B("a") where the intersection between these two sets may be significant. Thus, by creating a single output layer that has the total number of output units as the ordinal of the union of the corresponding bigram sets, we can map the corresponding input words to the same output layer. The input words that share the same output layer constitute an equivalence class. This notion is extended to an arbitrary number of input words with the dual goals of:

- Size of a given output layer should be as close to the size of the bigram set of its input word, $w_t$
- The total number of required output layers should be minimized by packing input words that have a large intersection of their bigrams in to a single output layer.

The above goals require us to find optimal lists of words that can be assigned the same output layer. The brute force approach of computing intersections of bigrams of all words and selecting the candidates for each output layer assignment is computationally prohibitive. We address this problem by modifying the 0/1 knapsack algorithm.

Another key consideration is to allow the language model to smooth the probability distribution for words that are not found in the bigram list of the training corpus for a word w but occurs in the test dataset. Traditional language models use some form of smoothing such as interpolation, discounting or back off techniques, each with their advantages and limitations. If our output layer for a given word w consists only of B(w) and a special word __unk__ to capture all words not seen in the training corpus, the ability of our model to smooth and generalize is severely restricted. In order to allow for a limited amount of generalization and also to minimize the total number of output layers, we introduce two mechanisms: (a) the notion of preset sizes, analogously termed as t-shirt sizes that are used to categorize B(w) and (b) integer factors, that along with t-shirt sizes determine the size of the output layers. The output layer for a given input word w along with its bigram list B(w) is determined as below. We first assign each B(w), based on its size, to one of the small number of t-shirt categories: tiny, small, medium, large, xlarge, xxlarge and ultra. For instance a word w which is a proper noun that has a follow list size of 10 is assigned to the tiny category. Common words in English such as "the" or "is" that are likely to have a large number of words that can follow them would get assigned to the size bucket termed ultra with their commensurate size. The common characteristic of a t-shirt category is that the size of an output layer belonging to this category is upper bounded by a threshold. In order to assign an output layer for a word w of a given t-shirt size category, we first allocate an initial output layer whose size is proportional to the t-shirt size. Concretely, the size of the initial output layer equals the product of t-shirt size and the proportionality constant, termed factor, which is a positive integer. Increasing the value of the factor has the effect of creating a larger output layer that can fit more number of input words. As the number of words that can be included in the output layer can be controlled by the factor, the probability distribution is computed over more number of words, thus allowing a limited form of smoothing. Once a list of input words are assigned the same output layer, we then optimize the output layer size by mapping each element of the output layer (which is a softmax unit) to a word belonging to the union of B(w) for each input word w that is assigned the same initial output layer. The resulting output layer is used as the softmax output layer of the RNN with suitable word mapping. Having segmented the input space of all words w in the vocabulary as per the size of the B(w), we address the problem of determining and packing words in to a shared output layer using the discrete 0/1 knapsack algorithm [17]. We treat the initial output layer as a knapsack bin that has a fixed maximum capacity. In our case this maximum capacity equals the size of the initial output layer, which is the product of its t-shirt size and the factor. The items that can be added to the knapsack are the words that belong to B(w) for a given w. We assign the cost of adding any word to the output layer to be 1 and the benefit offered by each word is also 1. Thus, the cost incurred in assigning an output layer to a given input word is the length of its bigrams B(w). This cost signifies the number of softmax units that are required in order to perform language model prediction for the given word. With these formulations that describe the output layer assignment in knapsack parlance, we outline the algorithm as shown in Algorithm A.1.

*D. Computational Complexity Analysis*

We observed from equations (1) and (2) that the factorization approach reduces the computation complexity by factorizing the output layer. The basis for complexity analysis for our technique is similar to the class based factorization described in [2] and characterized in equation (2), with a major difference. The class based approach involves a computation $H \times C$ per time step of the RNN and this term is constant if H and C are fixed. Our technique uses a variable sized output layer for every time step instead of classes. Hence the computations performed by the RNN per time step is proportional to:

$$O = (1 + H) \times H \times \tau + H \times E \quad (6)$$

where E is the expectation value of the size of the output layer. The definition and determination of E are as follows.

At every time step of the RNN computation, we choose the output layer based on the input word. As the size of the output layer is variable and the RNN goes through many time steps over a large number of input sequences, the computational complexity is expressed in terms of the expected value of the output layer size per time step. This is obtained by taking in to account the frequency of occurrence of the word tokens in the corpus with respect to the size of their respective output layers. The distribution of word tokens in the corpus with respect to the categories is illustrated in Fig 5 and Fig 6 for the Brown and the custom product review corpus.

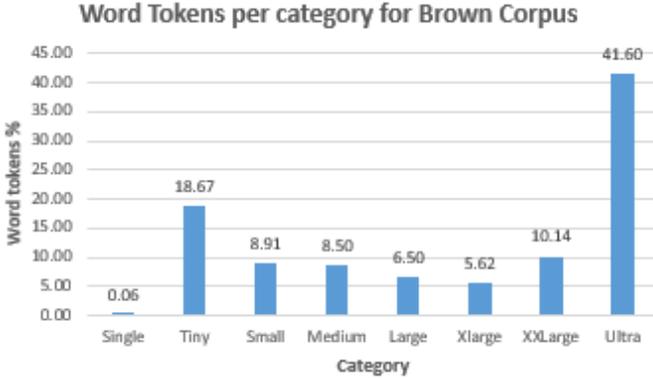

**Fig 5**: Categorywise distribution of words in Brown corpus

The distributions in Fig 5, Fig 6 imply that a very small subset of the vocabulary have the largest unigram counts and they also have the highest number of bigrams as in the Ultra category. We compute the expected value E of the size of an output layer per time step as below.

Let the size of a category $C_i$ be $S(C_i)$ and the probability of occurrence of a word w in $C_i$ in the training corpus be $p(C_i)$. Then: $\quad E = \sum_i S(C_i) p(C_i) \quad (7)$

Without the factorized model, the full softmax requires computations proportional to |V| for determining the output activations per time step. The improvement in efficiency can be measured as the ratio of |V| over E.

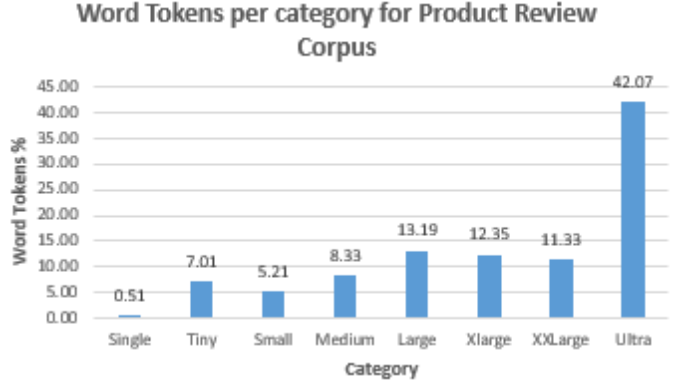

**Fig 6**: Categorywise distribution of words in custom corpus

### IV. IMPLEMENTATION

We implemented the RNN architecture shown in Fig 2 with tanh non linearity for the hidden layers with a factorized output layer performing softmax. The input words are mapped to their indices in the vocabulary and treated as a 15 bit binary vector. For efficiency purposes and also to filter noise in the corpus, we set a threshold of 5 on the unigram counts of the words. Any word whose unigram count falls below this threshold is treated as a special word: __unk__. These rare words may appear in the input or in the output. We assign a special binary vector for __unk__ at the input. For the output, we added this special word to each of the output layers so that it is predicted exactly in the same way as other words are predicted. Our text normalization also included case conversions where the corpus is converted to lowercase before further processing. The t-shirt size thresholds were chosen to be 32, 64, 128, 256, 512, 1024 and 2048 with variable factors for the different categories. The product of the factor and the t-shirt size determines the maximum size of the output layer in that category. Thus a word that has the size of its follow list up to (2048 * factor) can be fitted in to an output layer without omitting any word in its output vocabulary. If the bigrams of a word exceeds this size limit, we omit the least occurring bigrams. Though it is possible to use powerful variants of knapsack formulation, for the initial implementation we chose a single bin for simplicity.

### V. EXPERIMENTAL RESULTS

A good language model should assign high probability to sentences that are likely to occur in the language and low probability to those that are less likely. The most widely used metrics to evaluate language models are perplexity and word error rate (WER). Perplexity metric is quite popular as it allows easy comparison of different language models and is defined as:

$$Perplexity = 2^{-\sum_x p(x) \log_2 p(x)}$$

where x is a test sentence. As the goal of our work is to minimize the training time without increasing the perplexity compared to the other benchmarks, we evaluated our model on two core metrics: (a) Perplexity and (b) Training time. Our training data size in terms of word tokens and the vocabulary size are kept very similar to the earlier work in order to evaluate. Our work reported in this paper is primarily focused on optimizing the size of softmax layer with a perplexity comparable to or better than state of the art numbers. While the time taken to train the system is a direct measure of the speed improvements, it is hard to perform an apples to apples comparison with earlier work because factors like the specific hardware configuration, the libraries, operating system etc play a major role in determining the raw speed. Hence we report the output layer sizes used for computation as a key metric and also provide the wall clock time that helps a coarse level comparison. The language model architecture described in this paper was primarily developed to support applications in the domain of product reviews of mobile devices. Hence the evaluation was done primarily on this custom corpus. However for the purposes of benchmarking with other approaches we also evaluated the system on Brown corpus available with nltk 3.0 [15]. The Brown corpus with 57340 sentences was used for the evaluation. We used 40000 sentences for training with a token count of 888291 word tokens. With the minimum unigram count threshold of 5, the size of the vocabulary that we used has 14221 words. The size of the corpus and the vocabulary was chosen such that it is feasible to benchmark with Mikolov's architecture that also had a 5 count threshold and 800K word tokens. The training time reported in [2] was about 6 hours for this corpus and BLAS library [16] was used for speed up. In comparison, our implementation in Python using numpy takes about 90 mins per epoch of training and we found the model to yield best results within 1 to 3 epochs for the datasets we used. The expected value of the output layer size for the Brown corpus was determined to be 2111.35 without accounting for the intersections between the bigrams and the corresponding efficiency improvement compared to the full softmax was 6.73. After taking in to account the intersections during the generation of output layers, the effective expectation value of the size of the output layer is 445 for the Brown corpus, yielding an improvement of 31.9. The respective numbers for the custom corpus are: 2215.87 and 6.90 when intersections are not considered and the efficiency improvement is 23.35 when output layers are constructed taking in to account intersections between the bigrams. The expected value of the size of output layer is 655.41. An intuitive explanation for the difference in the expected value of output layer sizes between the Brown corpus and the custom corpus can be made by observing that the Brown corpus has a lot more words in the Tiny t-shirt category (18.67%) as compared to the custom corpus that has only 7.01% in the same category. This suggests that more number of output layers of Tiny category are generated for the Brown corpus, thus bringing down the expected value of output layer size. The model performance for the Brown corpus is tabulated in Fig 8. The max size shown in the table as in Fig 9 is the upper bound on the number of softmax units in a given category. The actual number of units in a given output layer is often less than the upper bound due to the intersection of elements between the different bigrams. We observed that the performance of our model scales well with other datasets, Fig 10 depicts the key metrics for the custom corpus, where the perplexity is lower and the training time gains are preserved.

| No | HUnits | Trg time (mins) | Trg set Size (tokens) | Test set size (tokens) | Perplexity |
|---|---|---|---|---|---|
| 1 | 16 | 72 | 888291 | 42046 | 177.91 |
| 2 | 16 | 72 | 888291 | 78050 | 210.75 |
| 3 | 32 | 91 | 888291 | 42046 | 168.41 |
| 4 | 32 | 91 | 888291 | 78050 | 184.63 |
| 5 | 48 | 100 | 888291 | 42046 | 143.4 |
| 6 | 48 | 100 | 888291 | 78050 | 175.53 |
| 7 | 64 | 125 | 888291 | 42046 | 227.88 |
| 8 | 64 | 125 | 888291 | 78050 | 243.85 |

**Fig 8**: Performance of the model on Brown Corpus

| Category | Number of output layers | Max Size (t-shirt_size * factor) |
|---|---|---|
| Tiny | 415 | 320 |
| Small | 113 | 512 |
| Medium | 10 | 640 |
| Large | 21 | 1024 |
| XLarge | 10 | 2048 |
| XXLarge | 8 | 3072 |
| Ultra | 8 | 6144 |

**Fig 9**: Number of output layers category-wise (Brown corpus)

| No | HUnits | Trg time (mins) | Trg set Size (tokens) | Test set size (tokens) | Perplexity |
|---|---|---|---|---|---|
| 1 | 16 | 40 | 499827 | 118604 | 148.41 |
| 2 | 32 | 61 | 499827 | 118604 | 140.59 |

**Fig 10**: Performance of the model on custom corpus

## VI. CONCLUSIONS AND FUTUTRE WORK

Our experiments showed that the RNN based language models using the factorization technique produce results that compare well with the benchmarks. We also observe that the performance is far superior on our custom corpus on product reviews. One possible reason for this is that the percentage of rare words is considerably smaller as many proper nouns (such as brand names, product names etc.) occur more regularly compared to the proper nouns encountered in the corpora used for benchmarking. The time for training the system is about 2 hours which is an improvement over other models that use similar sized corpora and vocabularies. A single output layer fits many words that have a sizable intersection of their bigrams enabling the model to generalize between similar words. One possible area for the future work is to enhance the performance by letting the model generalize better for the unknown words. We also intend to experiment with RNNs with other architectural variants such as LSTMs.

**Algorithm A.1** GenerateOutputLayers
Inputs:
    C – Output Layer Capacity
    W – Set of input words belonging to a t-shirt size category.
    B – A dictionary mapping for each word w in W to its respective bigram
Outputs:
    L – The list of output layers generated by this algorithm

$words \leftarrow W$
$L \leftarrow [\ ]$
$Benefits\_i = 1$ # we assign benefit for any word to be 1
$while\ words \neq \emptyset$
    $Initialize: Benefits = [\ ]$ # this is a 2 dimensional table
    $Initialize: Costs = [\ ]$
    $for\ i \leftarrow 0\ to\ length(words)$
        $Cost\_i = length(B(words[i]))$
        $Append\ Cost\_i\ to\ Costs$
        $for\ j \leftarrow 0\ to\ C$
            $if\ i = 0\ or\ j = 0$
                $Benefits[i][j] = 0$
            $else\ if\ Cost\_i > j$
                $Benefits[i][j] = Benefits[i-1][j]$
            $else$
                $if\ (Benefits[i-1][j-Cost\_i] + Benefits\_i) > Benefits[i-1][j]$
                    $Benefits[i][j] = Benefits[i-1][j-Cost_i] + Benefits\_i$
                $else$
                    $Benefits[i][j] = Benefits[i-1][j]$
    $i \leftarrow length(words)$
    $j \leftarrow C$
    $Initialize\ UnionBigrams \leftarrow \{\ \}$
    $Initialize\ SelectedWords \leftarrow \{\ \}$
    $while\ i > 0\ and\ j > 0$
        $if\ (Benefits[i][j] \neq Benefits[i-1][j]$
            $word \leftarrow words[i-1]$
            $add\ word\ to\ SelectedWords$
            $j \leftarrow j - Costs[i]$
            $UnionBigrams \leftarrow UnionBigrams\ \cup\ B(word)$
        $i \leftarrow i - 1$
    $words \leftarrow words - SelectedWords$
    $Append\ UnionBigrams\ to\ L$
$return\ L$